\documentclass[11pt,a4paper]{article}
\usepackage[hyperref]{emnlp-ijcnlp-2019}
\usepackage{times}
\usepackage{latexsym}

\usepackage{url}

\definecolor{light-blue}{rgb}{0.6,0.6,1}
\definecolor{orange}{rgb}{1,0.58,0	}

\usepackage{color}
\usepackage{booktabs}       
\usepackage{graphicx}
\usepackage{amssymb}
\usepackage{amsmath}

\usepackage{array}
\usepackage{multirow}
\newcommand{\ra}[1]{\renewcommand{\arraystretch}{#1}}

\def\bow{\textsc{GLOSS-BoW}}
\def\pos{\textsc{GLOSS-POS}}

\usepackage{hyperref}
\hypersetup{
    colorlinks=true,
    linkcolor=blue,
    filecolor=magenta,      
    urlcolor=blue,
}

\usepackage{float}

\usepackage{bm}
\usepackage{bbm}
\aclfinalcopy

\newcommand\glosspos{\textsc{GLOSS-Pos}}
\newcommand\glossbow{\textsc{GLOSS-BoW}}
\title{GLOSS: Generative Latent Optimization of Sentence Representations}

\author{Sidak Pal Singh \\
  EPFL\Thanks{ Work done while at Facebook AI Research} \\
  {\tt sidak.singh@epfl.ch} \\\And
  Angela Fan \\
  Facebook AI Research \\
  LORIA, Nancy \\
  {\tt angelafan@fb.com} \\\And
  Michael Auli \\
  Facebook AI Research \\
  {\tt michaelauli@fb.com} \\}
\date{}

\begin{document}
\maketitle
\begin{abstract}
We propose a method to learn unsupervised sentence representations in a non-compositional manner based on Generative Latent Optimization.
Our approach does not impose any assumptions on how words are to be combined into a sentence representation. 
We discuss a simple Bag of Words model as well as a variant that models word positions.
Both are trained to reconstruct the sentence based on a latent code and our model can be used to generate text.
Experiments show large improvements over the related Paragraph Vectors. 
Compared to uSIF, we achieve a relative improvement of 5\% when trained on the same data and our method performs competitively to Sent2vec while trained on 30 times less data.
\end{abstract}

\section{Introduction}

Learning sentence representations typically leverages the compositional structure of language~\cite{partee1984compositionality, Choi:2008:LCS:1613715.1613816}.
Common techniques include averaging word embeddings~\citep{pagliardini2017unsupervised, arora2017asimple} or distributions~\cite{singh2018context} as well as LSTM based encoders~\citep{kiros2015skipthought,hill2016learning} amongst many other approaches.

In this paper, we investigate a different direction where we seek to learn representations of sentences in a non-compositional manner.
We build on Generative Latent Optimization (GLO;~\citealp{bojanowski2018glo}) which learns a latent representation of each training example to be used by a decoder model that reconstructs the example.

GLO was previously applied to computer vision to disentangle the impact of convolutional neural networks versus adversarial training~\citep{goodfellowNIPS2014_5423} on image generation. 
\citet{bojanowski2018glo} show that good performance can be achieved without adversarial training and by freely optimizing a sample specific code via GLO (\textsection\ref{sec:glo}).

We apply GLO to learn sentence representations with an approach dubbed GLOSS. 
We jointly optimize a latent vector for each sentence and a decoder model that enables reconstruction of the sentences based on the latent vectors. 
This approach requires the model to learn how individual words are to be composed into a sentence representation. 

Our experiments on text similarity tasks show that GLOSS can achieve results which outperform state-of-the-art uSIF~\citep{ethayarajh} by 5\% when trained on the same amount of data.
It also performs competitively to another state-of-the-art method, Sent2vec~\citep{pagliardini2017unsupervised}, while being trained on an order of magnitude less data. 
The data efficiency of GLOSS training presents potential applications to low-resource languages.

\section{Related Work}
\label{sec:related}

Previous work on learning sentence representations broadly falls in two categories: unsupervised and supervised methods. 
Our focus is on unsupervised representation learning which does not require labeled data and often performs competitively on downstream tasks.

\paragraph{Unsupervised methods.} 
These approaches typically utilize a large unlabeled text corpus to learn word representations which are then composed into sentence representations. 
This could be as simple as using a bag-of-words averaging of Glove \citep{pennington2014glove} word embeddings trained on a corpus such as CommonCrawl, which we refer to as Glove-BoW. 
Methods such as Smooth Inverse Frequency (SIF; \citealp{arora2017asimple}) and unsupervised Smooth Inverse Frequency (uSIF; \citealp{ethayarajh}) build on this but instead carry out a weighted average of word embeddings and principal component removal. 
Sent2vec~\citep{pagliardini2017unsupervised} explicitly learns word embeddings such that their average works well as a sentence embedding.
Skip-thought~\citep{kiros2015skipthought} requires ordered training data and uses an LSTM-based encoder to build embeddings trained to reconstruct the surrounding sentences.

Paragraph vectors~\citep{le2014distributed} is the closest existing method to our approach in that it learns individual representations for each paragraph. 
They train representations to predict a random subset of words for a given sentence whereas GLOSS requires reconstructing the entire sentence from the latent vector.

We also normalize the latent vectors to be inside a Euclidean ball which is not the case for paragraph vectors and one of our models has the notion of order through positional embeddings.
These differences result in significantly better performance when training only on a fraction of data compared to paragraph vectors.

\paragraph{Supervised methods.} 
Methods requiring labels generally use less training data as they can be more data efficient due to the better training signal that can obtained from labeled data. 
Examples include: InferSent~\citep{conneau2017supervised} which uses labelled entailment pairs, GenSen~\citep{subramanian2018learning} utilizing supervision from multiple tasks, and ParaNMT~\citep{wieting2018paranmt} with paraphrase sentence pairs or conversational responses \cite{cer2018universal}.

\section{Generative Latent Optimization} 
\label{sec:glo}

The key idea behind Generative Latent Optimization~\citep{bojanowski2018glo} is to learn a latent vector $z \in \mathcal{Z} \subseteq \mathbb{R}^d$ for each data point $x \in \mathcal{X}$.
The latent vector $z$ is optimized to reconstruct $x$ with a decoder $g_\theta: \mathcal{Z} \to \mathcal{X}$. 
The objective is to \emph{jointly} optimize latent vectors $\{ z^1, \ldots, z^\textrm{N} \}$, corresponding to data points $\{ x^1, \ldots, x^\textrm{N} \}$, as well as the decoder parameters $\theta$ to minimize the empirical reconstruction loss:
$$
  \min_{\theta}~~\frac{1}{\textrm{N}}\sum_{i=1}^\textrm{N}~\left[~\min_{z^i \in\mathcal{Z}}~~\ell\left(g_\theta(z^i),x^i\right)\right]
$$
where $\ell$ is a differentiable loss function that measures the reconstruction
error and we can optimize it with stochastic gradient methods.
One challenge with GLO is that we have to learn a separate latent code for each data point.
In our setting we can use the latent codes as sentence representations.

\section{GLOSS}
\label{sec:gloss}

In this section, we apply GLO to learn sentence representations.
We detail the latent space $\mathcal{Z}$, the parameterization of the decoder, the training objective, and how we perform inference.

\subsection{Latent space}

In our setting, $\mathcal{X}$ denotes the space of sentences $\{ x^1, \ldots, x^\textrm{N} \}$ in a training corpus and $\mathcal{Z}$ denotes the space of possible sentence representations and we optimize a free vector $z \in \mathcal{Z}$ for each sentence.
As latent space we use the Euclidean ball with radius $r$~\citep{bojanowski2018glo}, i.e., $\mathcal{Z}=\mathcal{B}(r)= \left\lbrace z \in \mathbb{R}^d : \| z \|_2 \leq r \right\rbrace$. 
After every gradient step, we project the latent vectors back to $\mathcal{B}(r)$ with $\max(\|z\|_2, r)$. 
We initialize the latent vectors from a normal distribution over $\mathbb{R}^d$.

\subsection{Bag of Words Model}
\label{sec:bow}

The first model treats a sentence as a bag-of-words (BoW) which ignores the order of words.
The decoder is a linear transformation of the latent code $z^i$ to a vocabulary-sized vector to which we apply a sigmoid $\sigma$: $g_\theta(z) = \sigma(Wz + b)$ where $W \in \mathbb{R}^{V \times d}, b \in \mathbb{R}^V$ and $V$ is the vocabulary size.

Each output $o_j$ represents the probability of a particular word occurring in the current sentence. 
The target is a one-hot vector whose size is equal to the vocabulary and an entry is non-zero if word $w_j$ is present in the current sentence $s^i$, i.e., $(x^i)_j = \mathbbm{1}{\lbrace w_j \in s^i \rbrace}$, where $j \in \lbrace{1\ldots V\rbrace}$.
If the same word type occurs multiple times in a sentence, then we still predict it only once.
We train this model with a binary cross-entropy (BCE) loss to measure the reconstruction error between the outputs $o$ and target $t$: 
$$
\ell(o, t)  = - \sum_{j=1}^{V}\Big(t_j \log(o_j) + 
(1-t_j) \log(1-o_j)\Big)
$$

\subsection{Positional Model}

Bag-of-words models are very competitive in many NLP tasks, however, these models do not encode useful information about word order. 
For example, certain tasks require sentence representations to be able to differentiate between \textit{john eats a sandwich} and \textit{a sandwich eats john}. 
Therefore we consider a model which requires words to be reconstructed in their original order.

\begin{figure}
\centering
\includegraphics[width=0.4\textwidth]{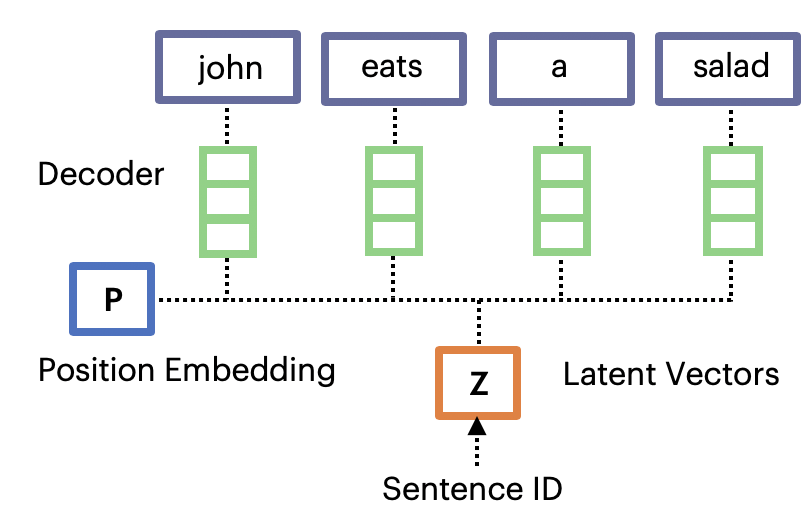}
\caption{Illustration of the \pos{} model.}
\label{fig:gloss-pos}
\end{figure}

\pos{} predicts each word in the sentence individually and each prediction is conditioned on the current word position (Figure \ref{fig:gloss-pos}).
For a sentence with $L$ words, we make $L$ predictions by adding a learned position embedding $p^l \in \mathbb{R}^d$ to the latent vector $z^i$ and then feed this to the decoder: $g_\theta(z^i + p^l)$.
The decoder is a simple linear transformation as in \textsection\ref{sec:bow}, except that we apply a softmax to the output of the linear projection and then apply a multi-class cross-entropy loss to each prediction.

\subsection{Inference}

We obtain sentence representation for unseen sentences by performing gradient descent over the input as follows:
we feed a randomly initialized $z$ to the decoder and minimize the reconstruction loss of the new sentence while keeping the decoder parameters $\theta$ fixed~\citep{bojanowski2018glo,le2014distributed}. 
In practice, we do this for 250 gradient steps with a learning rate of 1.

\subsection{Implementation details} 

We optimize models with Adam~\citep{Kingma2014AdamAM} with a learning rate of 0.0003, gradient-norm clipping of 25, and train for 210 epochs. 

We use $r=2$ as the radius of the Euclidean ball $\mathcal{B}(r)$. 
As training corpus we use a 2 million sentence subset of Newscrawl~\citep{bojar2018wmt} which amounts to 27M tokens.

\begin{table*}[!t] \centering\ra{1.1}
\small

\setlength{\tabcolsep}{4pt}
\begin{tabular}{@{}l|cc|cccccc|c|ccccc|c@{}}
\toprule
& \multicolumn{2}{c|}{Config.} & \multicolumn{7}{c|}{Unsupervised STS-() tasks} & \multicolumn{6}{c}{Supervised tasks}  \\
 			\cmidrule(lr){2-3} \cmidrule(lr){4-10} \cmidrule(lr){11-16}
Model  & \#Tok & Dim   & 12 & 13* & 14 & 15 & 16 & B & Avg & MR & CR & SUBJ & MPQA & TREC & Avg \\ 
\midrule

\multicolumn{6}{l}{\textit{Unsupervised methods trained on unordered corpus}}\\
\midrule
Glove-BoW $^\alpha$   & 840B & 300 & 52.2  & 49.6  & 54.6  & 56.3  & 51.4   &   41.5 & 50.9  & 77.3  & 78.3  & 91.2  & 87.9  & 83.0   &  83.5         \\
SIF       &  840B & 300 &  56.2 &  63.8     
&   68.5     &  71.7     & ---      &          72.0 & --- & --- &  ---      &  ---      &  ---     &   ---    &    ---      \\

uSIF     & 840B & 300 & \textbf{64.9}  &  \textbf{71.8}
&  \textbf{74.4}     &  \textbf{76.1 }     & ---      &      71.5 & --- & --- &  ---      &  ---      &  ---     &   ---    &    ---          \\
uSIF $^\alpha$   & 27M & 100 & 57.3 & 60.8  &  66.1  &  68.0   &  64.0        &     61.9 & 63.0 & --- & --- & --- & --- & --- & ---  \\
\midrule
PV-DBOW $^\beta$  & 0.9B & 300 & --- & ---  & 41.7 & ---  & ---   & 64.9  & --- & 60.2 & 66.9 & 76.3 & 70.7 & 59.4 & 66.7\\
Sent2vec    & 0.9B & 700 & 55.6 & 57.1 & 68.4 & 74.1 & 69.1 & 71.7 & 66.0 & 75.1	& 80.2& 	90.6& 86.3 &	83.8 &  83.2     \\
\midrule

\bow{}       &  27M & 100      & 54.8 & 51.8 & 68.4 & 71.2 & \textbf{71.8} & \textbf{72.4} & 65.1 &  67.4 &	72.0	& 86.4	& 79.5	&71.0 &   75.3   \\
\bow{}       & 27M & 300      & 55.9 & 55.6 &	69.2 & 	73.4 & 71.2 & 72.1 & 66.2 &  69.5	& 74.7	& 88.6	 & 82.3	& 78.0 &  78.6      \\

\pos{}    &  27M & 1K     & 53.0	& 52.9	& 67.5	& 72.0 & 69.8	& 68.0 & 63.9 &  73.4 &	78.1 &	91.0 & 86.3 &	82.1 &	82.2	
 \\
\midrule
\multicolumn{6}{l}{\textit{Unsupervised methods trained on ordered corpus}}\\
\midrule
 Skip-thought$^\dag$   & 0.9B & 2.4K  & 30.8 & 25.0 
&  31.4 & 31.0 & --- & --- & --- & 76.5 & 80.1 & \textbf{93.6} & 87.1 & \textbf{92.2} & 85.9 \\
\midrule
\multicolumn{6}{l}{\textit{Supervised methods trained on labeled corpus}}\\ 
\midrule

InferSent  (AllNLI) & 26M & 4.1K & 59.2 & 58.9 & 69.6  & 71.3 & 71.5 & 70.6  & \textbf{66.9} &   \textbf{81.1} & \textbf{86.3} & 92.4 & \textbf{90.2} & 88.2 & \textbf{87.6}\\
\bottomrule
\end{tabular}
 
\caption{\label{tbl:sts}Test performance on unsupervised STS-(12-16, B) and supervised tasks. 
($^\alpha$) indicate results computed by us. 
($^\beta$) PV-DBOW results are taken from ~\citet{pagliardini2017unsupervised}. 
($^\dag$) The unsupervised results for Skip-thought are taken from~\citet{arora2017asimple} and the supervised ones from ~\citet{pagliardini2017unsupervised}.
$(^*)$ Following SentEval, STS-13 does not include the SMT dataset due to licensing issues. The STS-B scores of other baselines are from \href{http://ixa2.si.ehu.es/stswiki/index.php/STSbenchmark}{the official webpage}, except for InferSent which are from~\citet{wieting2018paranmt}. All other results are from the respective publications.
See Table~\ref{tbl:sts-supp} in supplementary material for results with higher dimensionality.
}

\end{table*}

\section{Results}
\label{sec:results}

\subsection{Unsupervised tasks}

We first measure performance on the Semantic Textual Similarity (STS) tasks from SemEval 2012-2016 \citep{agirre2012sem, agirre2013sem, agirre2014sem, agirre2015sem, agirre2016sem} and STS-Benchmark (STS-B; \citealp{Cer_2017}). 
All tasks require the assignment of a similarity score to sentence pairs which are from domains such as Twitter, news headlines, online forums etc. 
Performance is measured in terms of Pearson correlation between the model score and human judgements and we report coefficients multiplied by 100. 
We tune hyper-parameters on the dev set of STS-B and report results on the test set of each benchmark.

We compare to a variety of unsupervised sentence embedding methods which are trained on different datasets and data sizes: Glove-BoW~\citep{pennington2014glove}, SIF~\citep{arora2017asimple}, uSIF~\citep{ethayarajh} on CommonCrawl (840B tokens); 
Paragraph Vectors (PV-DBOW; \citealp{le2014distributed}) on AP-news corpus (0.9B tokens); 
Skip-thought~\citep{kiros2015skipthought} and Sent2vec~\citep{pagliardini2017unsupervised} on BookCorpus (0.9B tokens). 

For comparison, we also show the performance of a supervised state-of-the-art method: InferSent~\citep{conneau2017supervised} trained on the labeled 
AllNLI (26M tokens) dataset.
We also compare to uSIF when trained on the same corpus as GLOSS to equalize the amount of data used (840B tokens v. 27M tokens in our setting).

Table~\ref{tbl:sts} shows that both \bow{} and \pos{} outperform uSIF on average when trained on the same corpus, a relative improvement of 5\%. 
\bow{} performs particularly well on STS-Benchmark where it outperforms all methods.
\bow{} is competitive to Sent2Vec, even though the latter was trained on 30 times more training data. 
In fact, it also matches the performance of InferSent which requires labeled training data.
This shows that GLOSS is very data efficient which makes it attractive for low-resource languages.

We observe that \bow{} is better than \pos{} for unsupervised tasks (cf. Table~\ref{tbl:sts-supp} in supplementary material).
Generally, increasing dimensionality does not improve accuracy on unsupervised tasks. 
This is in line with~\citet{Hill_2016} who observe that simple and shallow models work best for unsupervised tasks.

\subsection{Supervised tasks}

We evaluate model performance on several supervised classification tasks:
sentiment analysis (MR; \citealp{mrPang2005SeeingSE}), 
product reviews (CR; \citealp{crHu2004MiningAS}), 
subjectivity (SUBJ; \citealp{subjPang2004ASE}), 
opinion polarity (MPQA; \citealp{mpqaWiebe2005AnnotatingEO}), 
and question type (TREC; \citealp{Voorhees:2001:TQA:973890.973895}). 
Following 
SentEval, we train a logistic regression on top of the sentence vectors and report accuracy.

\begin{figure}
   \centering
   \includegraphics[width=0.4\textwidth]{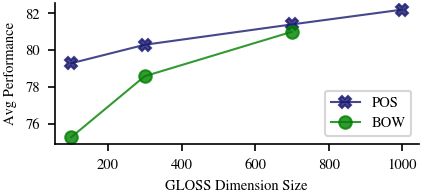}
 \caption{\bow{} and \pos{} average performance on supervised tasks for different $z$ dimension sizes.}
 \label{fig:supervised_curve}
\end{figure}

Figure~\ref{fig:supervised_curve} shows the comparative performance of \bow{} and \pos{} when increasing the dimension of $z$. \pos{} has stronger performance, but the simple \bow{} model improves with more dimensionality.

Table~\ref{tbl:sts} shows that InferSent performs best since it is trained on labeled data. 
Amongst unsupervised methods, \pos{} is slightly behind Sent2Vec which was trained on much more data. 
Both \pos{} and \bow{} outperform paragraph vectors (PV-DBOW).
Skip-thought and InferSent use much larger vectors and they require corpora with ordered sentences. 
We expect that increasing the dimensionality of $z$ even further would also help our models (cf. Figure~\ref{fig:supervised_curve}). However, we show that strong performance can be achieved with reasonable representation sizes.

\begin{table}[t]
\small
\centering
\begin{tabular}{@{}cl@{}}
\toprule
src & No comment on police suspension \\
(a) & Ryan unapologetic on stage suspension \\
(b) & Labour unapologetic the stage rule \\
(c) & Labour share the stage rule \\
(d) & Lock share us differently bread \\
tgt & History tells us differently . \\
\bottomrule
\end{tabular}
\caption{\label{tbl:interpolate} Generations (a, b, c, d) corresponding to latent space interpolations (on the Euclidean ball) between two sentences (src and tgt).}
\end{table}

\subsection{Text generation and interpolation}
\label{sec:textgen}

In contrast with previous methods, \pos{} enables text generation from sentence representations and can interpolate representations from two different examples.
Table \ref{tbl:interpolate} illustrates this for two sentences from the training set. Interpolations through the latent space show how words change going from the source to the target sentence. 

\section{Conclusion}

We apply Generative Latent Optimization to learn sentence representations in a non-compositional fashion.
Our models perform competitively to several well-known sentence representation methods, both on supervised and unsupervised tasks.
On unsupervised tasks, we outperform the popular uSIF method when trained on the same data.

\bibliography{master}
\bibliographystyle{acl_natbib}

\clearpage
\onecolumn

\section{Supplemental Material}
\label{sec:supplemental}
\begin{table*}[!ht] \centering\ra{1.1}
\small
\setlength{\tabcolsep}{4pt}
\begin{tabular}{@{}l|cc|cccccc|c|ccccc|c@{}}
\toprule
& \multicolumn{2}{c|}{Config.} & \multicolumn{7}{c|}{Unsupervised STS-() tasks} & \multicolumn{6}{c}{Supervised tasks}  \\
 			\cmidrule(lr){2-3} \cmidrule(lr){4-10} \cmidrule(lr){11-16}
Model  & \#Tok & Dim   & 12 & 13* & 14 & 15 & 16 & B & Avg & MR & CR & SUBJ & MPQA & TREC & Avg \\ 
\midrule

\bow{}       &  27M & 100      & 54.8 & 51.8 & 68.4 & 71.2 & \textbf{71.8} & \textbf{72.4} & 65.1 &  67.4 &	72.0	& 86.4	& 79.5	&71.0 &   75.3   \\
\bow{}       & 27M & 300      & \textbf{55.9} & 55.6 &	\textbf{69.2} & 	73.4 & 71.2 & 72.1 & \textbf{66.2} &  69.5	& 74.7	& 88.6	 & 82.3	& 78.0 &  78.6      \\
\bow{}         &  27M & 700      & 54.9 & \textbf{55.8}  &68.8 & \textbf{73.7} &71.0 &71.4  & 65.9 & 72.4	& 76.7	& 90.2	 & 83.7	& \textbf{82.2} &  81.0     \\
\midrule
\pos{}   &  27M & 100      & 54.6	& 54.8	& 68.3	& 71.7	& 71.4&	69.7  &  65.1 & 68.8 & 73.9 &	87.0	& 83.3 &	74.8	& 77.6 \\
 \pos{}    &  27M & 300      & 54.2 & 52.7	& 68.1	& 73.4 & 70.5 & 69.0 & 64.7 &  71.8 &	75.5	& 89.3	& 84.7 &	80.2 &	80.3	    \\
\pos{}   &  27M & 700      & 53.6	& 53.3	& 67.8	& 73.3 & 70.1	& 68.1 & 64.4 &  72.7 &	77.4 &	89.9 &85.4 &	81.4 &	81.4	\\
\pos{}    &  27M & 1K     & 53.0	& 52.9	& 67.5	& 72.0 & 69.8	& 68.0 & 63.9 &  \textbf{73.4 }&	\textbf{78.1} &	\textbf{91.0} & \textbf{86.3} &	82.1 &	\textbf{82.2}	
 \\

\midrule
\bottomrule
\end{tabular}
 
\caption{\label{tbl:sts-supp}\textbf{Effect of dimensionlity}: additional results for the test performance of \glossbow{} and \glosspos{} models with different dimensionality of the latent vectors on unsupervised STS-(12-16, B) and supervised tasks. 
$(^*)$ Following SentEval, STS-13 does not include the SMT dataset due to licensing issues. Best results for each task are in \textbf{bold}.
}

\end{table*}

\end{document}